# *CliMedBERT: A Pre-trained Language Model for Climate and Health-related Text*

**Babak Jalalzadeh Fard, Jesse E. Bell**
College of Public Health
University of Nebraska Medical Center
{babak.jfard, jesse.bell}@unmc.edu

**Sadid A. Hasan**
Microsoft AI
Cambridge, MA
sadidhasan@gmail.com

## Abstract

Climate change is threatening human health in unprecedented orders and many ways. These threats are expected to grow unless effective and evidence-based policies are developed and acted upon to minimize or eliminate them. Attaining such a task requires the highest degree of the flow of knowledge from science into policy. The multidisciplinary, location-specific, and vastness of published science makes it challenging to keep track of novel work in this area, as well as making the traditional knowledge synthesis methods inefficient in infusing science into policy. To this end, we consider developing multiple domain-specific language models (LMs) with different variations from Climate- and Health-related information, which can serve as a foundational step toward capturing available knowledge to enable solving different tasks, such as detecting similarities between climate- and health-related concepts, fact-checking, relation extraction, evidence of health effects to policy text generation, and more. To our knowledge, this is the first work that proposes developing multiple domain-specific language models for the considered domains. We will make the developed models, resources, and codebase available for the researchers.

## 1 Introduction

The World Health Organization (WHO) recognizes climate change as the most significant health threat to humans in the 21$^{st}$ century, which comes in many forms and is expected to cause approximately 250,000 additional deaths annually from 2030 to 2050 [1]. With timely and effective adaptation, many of these health risks can be reduced or avoided, but this requires comprehensive studies and policies that are multi-sectoral, multi-system, and collaborative at different scales [2]. Reaching that goal needs a thorough understanding and synthesis of the ever-growing research to inform policies [3]. While systematic reviews are the only commonly accepted tool for evidence synthesis [4], exponential growth in literature makes manual methods unattainable and biased toward more restrictive inclusion criteria [5]. To overcome these limitations, a recent study used different machine learning methods to map climate and health scientific literature [5]. Such comprehensive mappings can provide necessary inputs into global climate and health assessment, especially considering that bibliographic analysis of climate change policy documents shows that only 1.2% of climate change papers have at least one policy citation [6]. While such advancements in climate and health mapping can help distinguish gaps and focus areas, it becomes more challenging to capture the hidden knowledge in the ever-increasing literature. To this end, we propose to develop relevant multi-domain LMs that can serve as a foundation for various downstream NLP tasks such as identifying similarities between health and climate concepts, relationship extraction, summarization, classification, policy text generation, etc. [7]. To the best of our knowledge, this is the first work that proposes developing multiple domain specific LMs with different datasets and fine-tuning variations for the considered domains.

## 2 Related Work

Our proposed work is motivated by pre-trained LMs, such as BERT, which have shown considerable improvements in different downstream NLP tasks where they are fine-tuned on other domain-specific corpora [8]–[11]. Various LMs are available separately in our considered domains that are similarly fine-tuned on domain-specific corpora. For example, ClimateBERT is a domain-adapted pretraining of DistilROBERTA over 1.6 million paragraphs of climate-related texts from different resources such as news, scientific articles, and reports and fine-tuned over three downstream tasks with considerable improvements compared to the base model over climate-related NLP tasks [8], [12]. BioBERT [9] is another domain-specific LM pre-trained on biomedical corpora over BERT. Its largest corpora contain over 21B words from Wiki, Books, and scientific articles from PubMed and PMC. diseaseBERT is a domain-specific LM that can distinguish different aspects of diseases in a document [13]. It created the model through disease knowledge infusion training of BERT, ALBERT, and four pre-trained biomedical LMs (BlueBERT, ClinicalBERT, SciBERT, BioBERT). It showed improvements in three biomedical NLP tasks over their base models in all six models.

## 3 Methodology

### 3.1 Datasets

Our pre-training dataset contains scientific literature and policy documents in pdf format. We will also use a smaller and separate set of documents for downstream fine-tuning tasks. The pre-training corpora consist of the 16,078 scientific papers that are listed by Berrang-Ford et al. 2021 in their evidence synthesis study [5], and a total of 311 related policy documents that are listed in three studies [6], [14], [15]. We are also considering other Global or national policy documents such as IPCC AR6 Chapter 7 [2] and the US National Climate Assessment Chapter 14 on Health [16]. We will also consider searching in the Overton[1] database for more climate and health policy documents at national or regional levels. After collecting all policy documents, we will check and remove possible duplications. For downstream fine-tuning of the models, we will collect 100 scientific articles on climate- and health-related topics published after April 2020, parse them into paragraphs and label each paragraph accordingly. This will consider documents that were not part of our pre-training corpora. We are also considering using the CLIMATE-FEVER dataset: a set of 1,535 climate change claims labeled as Support, Refute, or Not enough information [17]. We will use a python script to clean and convert texts into machine-readable format for the considered datasets.

### 3.2 CliMedBERT language model

CliMedBERT will be a transformer-based LM pre-trained on the described corpus over two previously mentioned LMs: ClimateBERT and diseaseBERT (Figure 1). diseaseBERT is created in six variations that use different BERT models pre-trained on the disease corpora [13]. We will pre-train the six variations of diseaseBERT into six CliMedBERT$_D$ variations. CliMedBERT$_C$ will be pre-trained over ClimateBERT. Therefore, will present seven pre-trained models and seven base models in the final comparison tables of pre-training metrics and downstream tasks.

We will first compare our vocabulary of words from our corpus to the datasets used for pretraining ClimateBERT and MedBERT and estimate the percent overlap. Lower overlap values highlight the higher need for extended vocabulary; therefore, we can expect improvements in the final evaluations. In the next step, we will extend the vocabulary of our initial LMs by adding the most common tokens from our corpus to our LMs'. We will then train our two models using masked language modeling (MLM) with cross-entropy loss as cost function in predicting masked tokens. After this step, we will further fine-tune our models over the downstream dataset on three downstream tasks: Named Entity Recognition (NER), Text Classification, and Fact Checking [9],

---

[1] https://www.overton.io/

[18]. NER applies to the process of distinguishing domain-specific nouns in a text. We will consider entity-level precision, recall, and F1 score for the NER test as evaluation metrics. For text classification, we will classify CLIMATE-FEVER and our labeled paragraphs as Related or Not-Related to climate health effects. We will report and compare the cross-entropy loss and F1 score for all the models. For Fact-checking, we will separate health-related claims from CLIMATE-FEVER. We will also select and annotate paragraphs from our downstream corpus in the same way as CLIMATE-FEVER and use the same evaluation metrics as in the previous steps to compare and report the performance of our models.

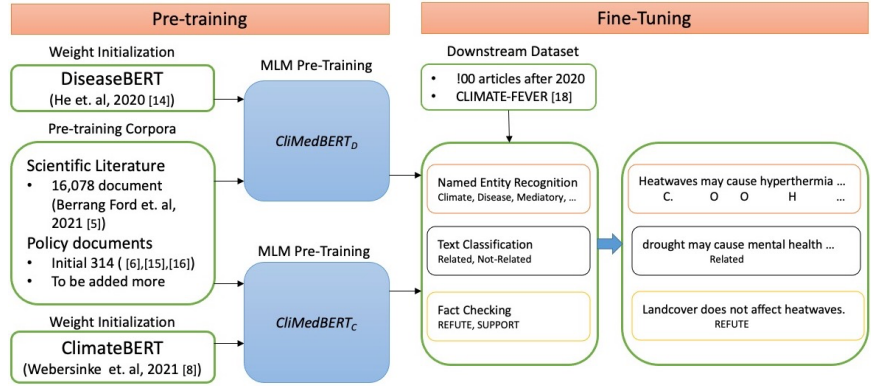

Figure 1: Overview of the pre-training and downstream fine-tuning of CliMedBERT.
*CliMedBERT$_D$ contains six LMs.

## 4 Expected Results

The main objective of this proposed work is to create a language model that best captures the relationships between climatic events, health outcomes, and the mediating factors from the corpus of scientific and policy literature. The language model can further capture the word sentiments on the health effects of climate change from the scientific literature and be used for different purposes or even find hidden relationships that have not been captured. We will provide the following outcomes from this study:

- Resulted LMs will be published as open access to be used for different purposes such as, but not limited to, evidence synthesis, evaluation of the comprehensiveness of published climate and health research in considering all factors and help to generation policy documents.
- Validation metrics and tests for the final LMs will be published.
- An article will be published containing the methodology and the results for different tests on the models. It will also analyze the potential effects of different factors, such as geographical location or developmental level, on the similarity and differences of the related health outcomes.
- A recreation of the results from Berrang. et al. 2021 [5] with our language model and comparing them to show the potential improvements.
- A classification engine for evaluating related texts and suggesting consideration of other variables based on factors such as the location or culture that can play a role in customized policy document generation based on specific health needs related to climate effects.

**Acknowledgments**

This work was supported by National Integrated Drought Information System (NIDIS) grant 13342-

Z7812001.